\documentclass[10pt,letterpaper]{article}
\usepackage[top=0.85in,left=1in,footskip=0.75in]{geometry}
\pdfoutput=1

\usepackage{amsmath,amssymb}
\DeclareMathOperator*{\argmax}{argmax}
\DeclareMathOperator*{\rank}{rank}

\usepackage{algorithm}
\usepackage[noend]{algpseudocode}

\usepackage{changepage}

\usepackage[utf8x]{inputenc}

\usepackage{textcomp,marvosym}
\usepackage{placeins}
\usepackage{float}
\usepackage{palatino}
\usepackage{booktabs}
\usepackage{makecell}
\usepackage{nonfloat}
\usepackage{multirow}

\usepackage[numbers]{natbib}

\usepackage{nameref,hyperref}
\usepackage{ccaption}

\usepackage{microtype}
\DisableLigatures[f]{encoding = *, family = * }

\usepackage[table]{xcolor}

\usepackage{array}

\newcolumntype{+}{!{\vrule width 2pt}}
\newlength\savedwidth

\def\ni{\noindent}
\usepackage{pdflscape}
\usepackage{afterpage}

\usepackage[aboveskip=1pt,labelfont=bf,labelsep=period,justification=justified,singlelinecheck=off]{caption}

\makeatletter
\renewcommand{\@biblabel}[1]{\quad#1.}
\makeatother

\date{}

\usepackage{lastpage,fancyhdr,graphicx}
\graphicspath{{figures/}}

\usepackage{epstopdf}
\fancyheadoffset[L]{0.25in}
\fancyfootoffset[L]{0.25in}

\begin{document}
\vspace*{-0.1in}

\begin{flushleft}
{\Large
\textbf\newline{Confusion-based rank similarity filters for computationally-efficient machine learning on high dimensional data} 
}
\newline
\\
KA Shapcott\textsuperscript{1,2*},
AD Bird\textsuperscript{1,2,3}
\\
\bigskip
\textbf{$^1$} Frankfurt Institute for Advanced Studies, Frankfurt-am-Main, 60438, Germany
\\
\textbf{$^2$} Ernst Str{\"u}ngmann Institute (ESI) for Neuroscience in cooperation with the Max Planck Society, Frankfurt-am-Main, 60528, Germany 
\\
\textbf{$^3$}  ICAR3R-Interdisciplinary Centre for 3Rs in Animal Research, Faculty of Medicine, Justus Liebig University, Giessen, 35390, Germany 
\\
\bigskip

%
%





*katharine.shapcott@esi-frankfurt.de

\end{flushleft}
\section*{Abstract}

We introduce a novel type of computationally efficient artificial neural network (ANN) called the rank similarity filter (RSF). RSFs can be used to both transform and classify nonlinearly separable datasets with many data points and dimensions. The weights of RSF are set using the rank orders of features in a data point, or optionally the `confusion' adjusted ranks between features (determined from their distributions in the dataset). The activation strength of a filter determines its similarity to other points in the dataset, a measure related to cosine similarity. The activation of many RSFs maps samples into a new nonlinear space suitable for linear classification (the rank similarity transform (RST)). We additionally used this method to create the nonlinear rank similarity classifier (RSC), which is a fast and accurate multiclass classifier, and the nonlinear rank similarity probabilistic classifier (RSPC), which is an extension to the multilabel case. We evaluated the classifiers on multiple datasets and RSC was competitive with existing classifiers but with superior computational efficiency. Open-source code for RST, RSC and RSPC was written in Python using the popular scikit-learn framework to make it easily accessible. In future extensions the algorithm can be applied to specialised hardware suitable for the parallelization of an ANN (GPU) and a Spiking Neural Network (neuromorphic computing) with corresponding performance gains. This makes RSF a promising solution to the problem of efficient analysis of nonlinearly separable data.

\subsection*{Keywords}
Machine learning; vector quantization; rank; confusion; nonlinear

\clearpage


\section*{Introduction}

Data that varies in a nonlinear manner is common in real world datasets, requiring nonlinear classifiers to separate classes. However, nonlinear classifiers are comparatively either computationally inefficient on these large datasets or need multiple runs to find appropriate parameters, for example choosing the correct kernel for a support vector machine (SVM). This inefficiency wastes energy, resources and time, especially as datasets have increased in size and dimensionality in recent years \cite{Ahmed2017-qv}. The novel method we present here was motivated by the desire to create an efficient, nonlinear classifier that can scale to big data and high dimensions with little prior knowledge of the data.

\vspace{4mm}

\ni To solve pattern recognition tasks like classification, vector quantization (VQ) methods have been widely used. These classifiers include SVM, learning vector quantization (LVQ), neural gas and self-organizing maps (SOM) and have been used to solve a wide range of machine learning tasks \cite{Villmann2017-ap}. Those that use vector prototypes have the additional benefits that they are intuitive, have a low complexity and are relatively computationally efficient. It has recently been shown that vector prototype layers can be added to deep neural networks in order to increase their interpretability while maintaining performance \cite{Li2018-ij}. Due to the increasing use of machine learning in society, moving towards intuitive models is an important goal \cite{Rudin2019-wv}.

\vspace{4mm}

\ni In this paper we present a new class of artificial neural network (ANN), rank similarity filters (RSF). These are filters which can perform supervised VQ using a weight matrix based on the order statistics of the input features. The activation strength of many of these filters in response to an input give a relative measure of similarity. We show that the order statistics of features are usually sufficient to approximate confusion values derived from the data. We created an open source package which implements the algorithm in Python using the scikit-learn framework (found here: \url{https://github.com/KatharineShapcott/rank-similarity}) and validate the efficiency and accuracy of our method on real-world datasets. We demonstrate that RSFs are computationally efficient in use of CPU time (and energy) during both training and evaluation. We show their utility in both data transform and classification and discuss possible extensions.

\vspace{4mm}

\ni This paper is organised as follows; first we demonstrate the novelty of RSF in comparison to related methods, next we justify our approach theoretically, then explain the rank similarity transform and classifier algorithms and finally we demonstrate the algorithms' computational efficiency compared to other methods as measured by elapsed CPU time.


\section*{Related work}


Our method is a novel type of nonparametric vector quantization (VQ). VQ is used not only for pattern recognition but also for data compression or approximate nearest neighbor searches \cite{Wu2019-so}. Since data are not spread evenly through the entire feature space, mapping the feature space into an appropriately spaced "codebook" results in a good approximation of the true data space at a much lower storage (and search) cost. K-means (or k-means++) is the most well known VQ method, and is popular because it is a fast and simple algorithm with many uses \cite{Hastie2009-tp}. K-means makes unsupervised partitions of the data into equal variance clusters and the centroids of those clusters make up the codebook. In VQ methods new datapoints can be assigned to part of the codebook according to their similarity.

\vspace{4mm}

\ni Biologically inspired VQ methods also exist, of which SOM is a well known example. SOM is a type of ANN with a topological and competitive learning algorithm that performs VQ \cite{Kohonen1982-sl}. It has been extensively used for nonlinear dimensionailty reduction and text classification and has even been modified to solve the travelling salesman problem \cite{Kohonen2013-hf}. Very briefly, when the SOM codebook is trained on a new data point the neuron (node) with the most similar weight vector (usually measured either with Euclidean distance or with a dot product) updates its weights towards the new vector. In the present work we also use the dot product as similarity measure due to its biological plausibility \cite{Koch1992_ka}. All other neurons then update their weights depending on their topological distance from that neuron. The neural gas \cite{Martinetz1991-ob} is a type of SOM which performs pure VQ. It does not update weight vectors according to a topography but instead according to their ranked similarity to the new training data point. 

\vspace{4mm}

\ni Supervised biologically inspired VQ methods can perform classification. SVM is a popular example and it is an extremely successful classifier across many datasets \cite{Fernandez-Delgado2014-ov}. However, it has the drawback that it is not intuitively understandable, since the "support vectors" in the codebook are the extreme borders between classes. LVQ is an ANN method which instead finds prototypical vectors which are similar to members of the class \cite{Kohonen1988-ci}. The generalized LVQ (GLVQ) is a mathematically tractable version in which prototypes update their weights to minimize a cost function \cite{Sato1995-aq}. Like SVM their performance can be improved with a kernel, however this may not be desirable as it makes it no longer intuitive \cite{Nova2014-va}. All these prototype VQ methods have the drawback of being computationally inefficient on modern high dimensional data sets. 
 
\vspace{4mm}

\ni In order to make an efficient codebook other VQ methods use e.g. sparseness or tree structures \cite{Wu2019-so}. Here we instead created a novel nonparametric codebook from the rank transform of datapoints, resulting in ranked prototypical vectors (the rank similarity filters - RSFs). Rank transformed data are robust to outliers, are agnostic to changes in scale and are nonparametric, sharing similar advantages to nonparametric statistical methods \cite{Conover1981-us}. As the codebook is based on ranks it need only store integer values up to the number of features $f$, which is memory efficient and creates a greatly reduced search space. Importantly, the prototypes in this codebook do not represent feature magnitude but instead reflect the relationship between their features, as detailed in the following section. 


\section*{Discriminability of ranked features}

\subsection*{Confusion and discriminability}

\begin{figure} [p] 
\centering
\hspace*{0cm}
\captionsetup{width=17cm}
\includegraphics[trim=0cm 0cm 0cm 0cm,clip,width=16cm]{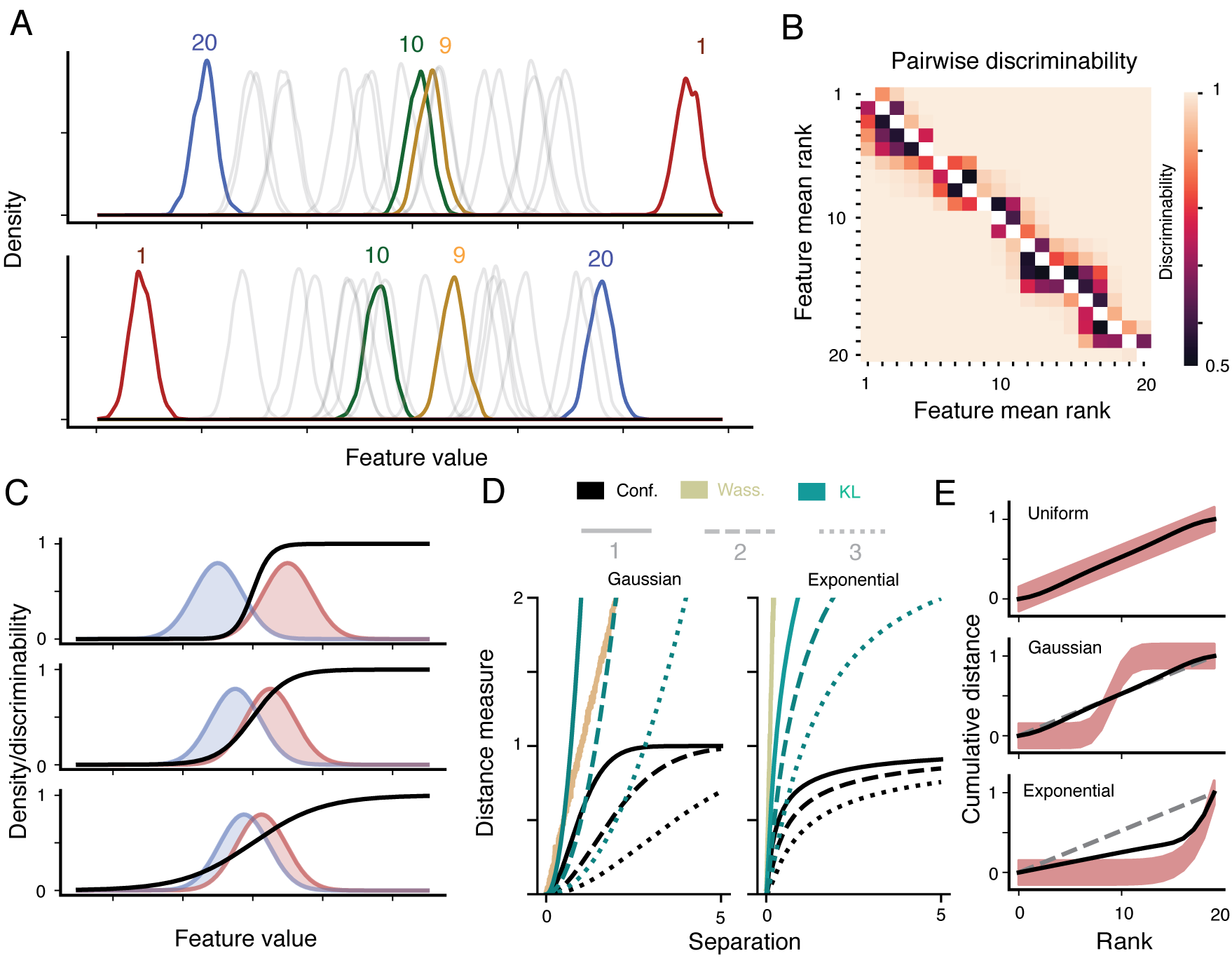}
\caption{\small \textbf{Separation of features using confusion. A}. Examples of the distributions of features of two classes (top and bottom) with $20$ features. The features are rank-ordered by mean for the top class (coloured lines) and the overlapping distributions show the potential overlap of different realisations. \textbf{B}. Pairwise discriminability between features values in each of the classes in Panel A. Top right corresponds to the top class and bottom left to the bottom class. \textbf{C}. Discriminability as a function of feature value (black line) for pairs of distributions with different degrees of overlap (top to bottom). Distributions with more overlap lead to a slower increase in discriminability with feature value. \textbf{D}. Comparison of different distribution distance measures as a function of the mean separation of two gaussians (lefts) and two exponential distributions (right), each with standard deviations of $1$ (solid lines), $2$ (dashed lines), and $3$ (dotted lines). Black shows the confusion measure, yellow the Wasserstein distance (same for all standard deviations), and green the Kullback-Liebler divergence. \textbf{E}. Example of the confusion measure (Eq \ref{eq:actualconfusion}, black solid line) applied cumulatively to features with means drawn from uniform (top), gaussian (middle), and exponential (bottom distributions). The red shaded region shows the variability around the mean of each feature and the grey dashed line shows the cumulative rank value (without accounting for confusion).}
\label{fig:confusion}
\end{figure}

\ni Classes in a dataset can be represented by estimating the distribution of the values taken by each of their features (Figure \ref{fig:confusion}A). If these features are ordered by their mean value (highlighted curves in Figure \ref{fig:confusion}A), it is possible to estimate the confusion between features: the probability that features taking certain values are drawn from one of two neighbouring distributions and not the other. In particular, given two distributions $f$ and $g$, the probability $\mathrm{P}_f(x)$ that an observed value $x$ is drawn from $f$ is
\begin{equation}
\mathrm{P}_f(x)=\frac{f(x)}{f(x)+g(x)}
\end{equation}
Taking expectation over $g$ gives a symmetric measure of confusion $c_{fg}$
\begin{equation}
c_{fg}=\int_{\mathrm{supp}(f)\cap\mathrm{supp}(g)}\frac{f(x)g(x)}{f(x)+g(x)}dx
\end{equation}
Where $\mathrm{supp}(f)$ and $\mathrm{supp}(f)$ are the supports of $f$ and $g$ respectively. If $f$ and $g$ are entirely disjoint the confusion will be $0$, and if $f \equiv g$ the confusion will take a maximum value of $0.5$.
\vspace{4mm}

\ni The discriminability of two distributions $f$ and $g$ can be defined as $d_{f,g}=1-c_{f,g}$ and is the expected probability that a given random sample can be assigned to the correct distribution based on its value. The minimum value of $d_{f,g}$ is $0.5$, which is obtained when $f\equiv g$. In this case there is still an even chance of assigning any sample to its correct distribution. This means that $d_{f,g}$ is not a metric on probability distributions. Figure \ref{fig:confusion}B plots the pairwise discriminability between each pair of features in the classes shown in Figure \ref{fig:confusion}A. The greater the overlap between distributions the lower the discriminability (Figure \ref{fig:confusion}C).

\subsubsection*{Comparison of confusion with other probability distances}
\ni A number of other measures can be used to quantify the distance between two probability distributions. The most widely-known are the Kullback-Liebler divergence \cite{Kullback1951_is} and the Wasserstein distance \cite{Dobrushin1970_wd}. Briefly, the Kullback-Liebler divergence $\mathrm{D_{KL}(f,g)}$ gives the information loss in representing samples from a distribution $f$ by distribution $g$ and the Wasserstein distance gives the weighted difference in probability between $f$ and $g$. Both are unbounded, with the Kullback-Liebler divergence being infinite for non-overlapping distributions and the Wasserstein distance growing linearly with the distance between distributions. In contrast the confusion measure $c_{f,g}$ reaches its finite maximum value when $f$ and $g$ are perfectly discriminable; when presented with a value drawn from either distribution it is possible to assign it to its source with $100\%$ chance. Figure \ref{fig:confusion}D shows this effect in action by comparing the distance measures for pairs of gaussian (left) and exponential (right) distributions with varying standard deviations as a function of the distance between their means. The Wasserstein distance (yellow) is independent of the standard deviations of the features around their means, and both the Kullback-Liebler divergence (green) and the Wasserstein distance grow unboundedly as a function of separation.

\subsection*{Mapping confusion to filters}
\ni The order statistics of features convey information about the identity of an object. We propose that the relative weighting of order statistics should depend on the confusion between their values over different examples in the training set. In particular, the size of a gap between values assigned to features with consecutive order statistics should depend on their discriminability, both from each other and the extreme values of the object. If $n$ features are sorted by their mean values over a representative training set, then their values are assumed to take distributions $\{f_1,f_2,...,f_n\}$, where the index represents the order of the sample mean. Then the map between consecutive values is given by
\begin{equation}
    M_{i,i+1}=\frac{1}{3}\bigg[d_{f_{i},f_{i+1}}+\bigg(d_{f_{1},f_{i}}-d_{f_{1},f_{i+1}} \bigg)+\bigg(d_{f_{n},f_{i}}-d_{f_{n},f_{i+1}}\bigg)\bigg]
    \label{eq:actualconfusion}
\end{equation}
In Figure \ref{fig:confusion}E, this process is applied to data generated with random feature means drawn from uniform (top), gaussian (middle), and exponential (bottom) distributions. In each case the distribution about the mean is gaussian (red shaded area shows two standard deviations around the mean). For data drawn from distributions without a high degree of skew (top two panels), the confusion measure is very well approximated by the mean rank (grey dashed lines). We find that this is the case in the majority of common datasets (with some exceptions, see below), and so propose the simpler rank-based filter as an efficient and powerful heuristic in most cases.

\vspace{4mm}


\section*{Rank similarity filters}

\subsection*{The algorithm}\label{methods:algorithm}
A short description of the training algorithm is as follows:
\begin{enumerate}
	\item OPTIONAL: A distribution for the filters is calculated
	\item Filters are initialized with random data points from the training data
	\item Filters are assigned to the data points that they are most strongly activated by.
	\item A new filter is created from the assigned data points. Repeat from step 3 until only $tol$ points move to another filter. 
	\item L1 normalize each filter using $x_{norm} = \frac{x}{\sum{x}}$.
\end{enumerate}
\vspace{4mm}

\ni This is shown in pseudo-code in \ref{alg:rf}.  
$X$ is the training data with $n$ features and $N$ samples. A single rank filter $f$ belongs to $F$ containing $m$ filters where $m$ is less than $N$. For features $i=1,...,n$ and for filters $j=1,...,m$.
\vspace{4mm}

\begin{algorithm}
  \caption{Rank Similarity Filters($X$) Creates filters $F$ from a dataset $X$}\label{alg:rf}
  \begin{algorithmic}[1]
    \Require{$X$} 
    \Ensure{$F$ (filters)}
    \Function{RankSimilarityFilter}{$X$, n-filters, distribution}
      \State $D  \leftarrow $ \Call{CreateDistribution}{$X$, distribution}
      \State $F  \leftarrow $ \Call{InitializeFilters}{$X$, $D$, n-filters}
      \State $F  \leftarrow $ \Call{SpreadFilters}{$X$, $F$, $D$}
      \State $F  \leftarrow $ \Call{L1norm}{$F$}
      \State \textbf{return} $F$ \EndFunction
  \end{algorithmic}
\end{algorithm}

\ni The first step of creating a distribution $D$ is optional because for many datasets the numeric ranks are already discriminable enough for practical use. A distribution can be created by any function that produces a sorted vector with a length equal to $n$. Creating the confusion distribution was performed as outlined in the Methods.
\vspace{4mm}

\ni For the initialisation of each filter a random data point is drawn from $X$ and the numeric rank calculated. This can either be used to draw from a distribution or returned directly as shown in Algorithm \ref{alg:init}. 

\begin{algorithm}
  \caption{InitializeFilters($X$, $D$, n-filters) Creates filters $F$ randomly from $X$ with values using distribution $D$}\label{alg:init}
  \begin{algorithmic}[1]
    \Require{$X$, $D$, n-filters} 
    \Ensure{$F$}
    \Function{InitializeFilters}{$X$, $D$, n-filters}
      \For{$i \gets 1,$n-filters}
        \State $x \gets $ \textbf{random}(X) \Comment{Without replacement}
        \State $r \gets \textbf{numeric-rank}(x)$
        \If{$D$ \textbf{is} \texttt{None} }
            \State $F[i] \gets r$
        \Else
            \State $F[i] \gets D[r]$
        \EndIf
      \EndFor
      \State \textbf{return} $F$ \EndFunction
  \end{algorithmic}
\end{algorithm}

\vspace{4mm}
\ni The initialized filters are then spread throughout the data using a greedy algorithm. The aim is to find the set of points $C_j$ in $X$ that are most similar to each filter $A_j$. To this end we find the maximum of a dot product across all filters for each $x$ in $X$
\begin{equation}  \label{eq:max}
    j = \argmax(A \cdot x)
\end{equation}
\vspace{4mm}

\ni To update the weights of filter $A_j$ we then compute
\begin{equation} \label{eq:rank}
    A_j = \rank(\overline{C_j})
\end{equation}

\ni Which, as before, can be used to optionally draw from distribution $D$ or not. This process is repeated until only $tol$ points move to a different filter in order to ensure the filters are spread evenly throughout the data, as shown in algorithm \ref{alg:spread}. 
\vspace{4mm}

\begin{algorithm}
  \caption{SpreadFilters($X$, $D$, $F$, $tol$) Spread filters $F$ throughout $X$ using distribution $D$ until only $tol$ change membership}\label{alg:spread}
  \begin{algorithmic}[1]
    \Require{$X$, $D$, $F$, $tol$} 
    \Ensure{$F$}
    \Function{SpreadFilters}{$X$, $D$, $F$, $tol$}
      \State $c \gets tol+1$
      \While{$c > tol$}
          \State $I \gets \textbf{argmax}(X \cdot F)$
          \ForAll{$f \in F$}
            \State \texttt{/* Update filters using most similar data points in X */}
            \State $m \gets \textbf{mean}(X[I = f])$
            \State $r \gets \textbf{numeric-rank}(m)$
            
            \If{$D$ \textbf{is} \texttt{None} }
                \State $f \gets r$
            \Else
                \State $f \gets D[r]$
            \EndIf
          \EndFor
          \State \texttt{/* Check if filters have converged */}
          \If{\textbf{exists}$(I_{prev})$ }
            \State $c \gets \textbf{count-nonzero}(I - I_{prev})$
          \EndIf
          \State $I_{prev} \gets I$
      \EndWhile
    \State \textbf{return} $F$ \EndFunction
  \end{algorithmic}
\end{algorithm}

\ni After performing the L1 norm the filters have been trained. This is enough to transform data points into a new space in an unsupervised manner based on their similarity $S$ to the filters $F$ (rank similarity transform). This can be simply done by performing a dot product of the dataset with the filters and then scaling:
\begin{equation}
\begin{split}
    S  &= F \cdot X \\
    S' &= (\frac{S_{max} - S_{min}}{S - S_{min}})
\end{split}
\end{equation}
\ni However, due to the "curse of dimensionality" only the largest values are informative about the filter similarity while other smaller values are less so \cite{Houle2010-ez}. We therefore chose a new informative minimum $S_{Imin}$ which is the activation of the $k$th most similar filter (where $k$ is a hyperparameter of the model, see Algorithm \ref{alg:rft}) and set all values less than this to zero. 
\begin{equation}
\begin{split}
    S^I &= (\frac{S_{max} - S_{Imin}}{S - S_{Imin}}) \\
    S^I_x &= 
    \begin{cases}
        S^I_x & S^I_x >0 \\
        0 & S^I_x<0
    \end{cases}
\end{split}
\end{equation}

\begin{algorithm}
  \caption{RankSimilarityTransform($F$, $X$) Transform dataset $X$ using filters $F$}\label{alg:rft}
  \begin{algorithmic}[1]
    \Require{$X$} 
    \Ensure{$S^I$}
    \Function{RankSimilarityTransform}{$F$,$X$,$k=25$}
      \State $S \gets X \cdot F$
      \ForAll{$s \in S$}
        \State $s_{max} \gets $\textbf{max}($s$)
        \State $s_{Imin} \gets $\textbf{partition}($k$,$s$)
        \State $s^I \gets (s_{max} - s_{Imin})/(s - s_{Imin})$
        \State $s^I \gets $\textbf{clip}($s^I$, 0, 1)
      \EndFor
      \State \textbf{return} $S^I$ \EndFunction
  \end{algorithmic}
\end{algorithm}

\subsection*{Classification algorithm}

In order to create a classifier from the rank similarity filters few additional steps in the algorithm are required. To perform classification each filter is assigned a label $l$ depending on the class makeup of its data points. For simple multiclass classification problems (i.e. single labels with multiple classes) performed by rank similarity classifier it is sufficient to separate $X$ into subsets $X_l$ per label and create the filters $F_l$.
This adds two additional steps (see Algorithm \ref{alg:rfc}), one before the main algorithm and one after:
\begin{enumerate}
	\item Split the data according to class.
	\item Assign each filter a label based on which class of data it was trained on.
\end{enumerate}

\begin{algorithm}
  \caption{Rank Similarity Classifier($X$) Creates filters $F$ with labels $l$}\label{alg:rfc}
  \begin{algorithmic}[1]
    \Require{$X$, $y$} 
    \Ensure{$F$, $l$}
    \Function{RankSimilarityClassifier}{$X$,$y$}
      \State $C  \leftarrow $ \textbf{unique}($y$)
      \ForAll{$c \in C $}
        \State $F[c]  \leftarrow $ \Call{RankSimilarityFilter}{$X[c]$}
        \State $l[c]  \leftarrow $ $c$
      \EndFor
      \State \textbf{return} $F$, $l$ \EndFunction
  \end{algorithmic}
\end{algorithm}

\vspace{4mm}

\ni Then to predict the label of a data point $x_l$ we find the maximally similar filter $j$ using \eqref{eq:max} then find the class $c$ of the data that $j$ was trained on and assign $x_l = c$. 
\vspace{4mm}

\ni For the probabilistic classifier the data does not need to be split but after training the label must be calculated:
\begin{enumerate}
	\item Assign each filter a label based on the labels of its data points weighted by activation.
\end{enumerate}

\vspace{4mm}

\begin{algorithm}
  \caption{Rank Similarity Probabilistic Classifier($X$) Creates filters $F$ with labels $l$}\label{alg:rfp}
  \begin{algorithmic}[1]
    \Require{$X$, $y$} 
    \Ensure{$F$, $l$}
    \Function{RSPClassifier}{$X$,$y$}
      \State $F  \leftarrow $ \Call{RankSimilarityFilter}{$X$}
      \State $l  \leftarrow $ \Call{SetLabel}{$X$, $y$, $F$}
      \State \textbf{return} $F$, $l$ \EndFunction
  \end{algorithmic}
\end{algorithm}

\ni To perform probabilistic classification (suitable for multiclass-multilabel problems) for filters $F$ the probabilistic label $L$ is calculated based on the normalized sum of the class identities of the maximally similar data points as calculated in (\ref{eq:max}). 

\vspace{4mm}

\begin{algorithm}
  \caption{SetLabel($X$, $y$, $F$) Creates probabilistic label matrix $L$ using the class identity $y$ of dataset $X$}\label{alg:label}
  \begin{algorithmic}[1]
    \Require{$X$, $y$, $F$} 
    \Ensure{$l$}
    \Function{SetLabel}{$X$, $y$, $F$}
      \State $I \gets \textbf{argmax}(X \cdot F)$
      \ForAll{$f \in F$}
        \State \texttt{/* Label filters using most similar data points in X */}
        \State $L[f] \gets $\Call{L1norm}{\textbf{sum}$(y[I = f])$}
      \EndFor
      \State \textbf{return} $L$ \EndFunction
  \end{algorithmic}
\end{algorithm}

\ni Then to perform a probabilistic prediction of the label of a data point $x_l$ we perform \verb+RankSimilarityTransform+  \ref{alg:rft} and use these values to assign probabilities to the classes. The maximum value is then the class label.

\vspace{4mm}

\begin{algorithm}
  \caption{Probabilities($F$, $L$, $X$) Creates class probabilities for dataset $X$ from filters $F$ with label $L$}\label{alg:prob}
  \begin{algorithmic}[1]
    \Require{$F$, $L$, $X$} 
    \Ensure{$P$}
    \Function{Probabilities}{$F$, $L$, $X$}
      \State $S \gets$ \Call{RankSimilarityTransform}{$F$, $X$}
      \ForAll{$s \in S$}
        \ForAll{$c \in C$}
          \State $p[c] \gets $\textbf{max}($L[s,c] * s$)
          \vspace{1mm}
        \EndFor
        $P[s]  \gets $ \Call{L1norm}{$p$}
      \EndFor
      \State \textbf{return} $P$ \EndFunction
  \end{algorithmic}
\end{algorithm}

\ni This \verb+Probabilities+ algorithm (\ref{alg:prob}) can be similarly used to calculate probabilities for the standard RSC but here the label $l$ for each class can take only the value $[1,0]$. 

\vspace{4mm}

\subsection*{Determining the number of filters}

The number of filters $m$ can be set directly as a parameter or calculated based on the number of training data points $N$ and two thresholds, $m_{min}=1000$ and $m_{max}=10000$. 
\begin{equation*}
\begin{aligned}
    m &= N, &&\text{ if  $N < m_{min}$} \\
    m &= m_{min}, &&\text{ if  $N < m_{min}*10$} \\
    m &= N/10, &&\text{ if $N < m_{max}$ } \\
    m &= m_{max}, &&\text{ if $N < m_{max}*10$ } \\
\end{aligned}
\end{equation*}

\ni The threshold of $m_{max}$ can be set to give the most efficient results based on the available computing resources.



\section*{Experimental Results}


In order to compare the efficiency of our proposed method to other classifiers we implemented the above algorithms in Python using the scikit-learn framework. The code is publicly available\footnote{\url{https://github.com/KatharineShapcott/rank-similarity}}. We evaluate the code on real-world datasets using separate SLURM jobs via the ACME package \cite{Fuertinger2021}. Each was allocated a single core of a CPU (Intel Xeon E5-2650 v2 or v3) and 8 GB RAM using Red Hat Enterprise Linux 8.1. Efficiency was measured by the CPU time used using the Python module \texttt{time}. Accuracy was evaluated using the unweighted mean of the F1 score.

\subsection*{Datasets}

We used three multiclass real world datasets to assess the classifiers. The first was Fashion-MNIST \footnote{\url{https://github.com/zalandoresearch/fashion-mnist}}, a more difficult and nonlinear version of the digit MNIST dataset. Fashion-MNIST is an image dataset with 10 classes (c), dimensionality (d) of 784 and consists of 60000 training samples and 10000 balanced test samples. Kuzushiji-49\footnote{\url{https://github.com/rois-codh/kmnist}} \cite{Clanuwat2018-ye} (d=784, c=49, 232,365 training samples and 38,547 test samples) and 20 Newsgroups\footnote{\url{http://qwone.com/~jason/20Newsgroups/}} (d=101,631, c=20, 11,314 training samples and 7,532 test samples) were also used. A fourth multiclass and multilabel dataset Reuters Corpus Volume 1 (RCV1)\footnote{\url{https://jmlr.csail.mit.edu/papers/volume5/lewis04a/}} \cite{Lewis2004-sg} dataset (d=47236 sparse, c=101) was used to assess the multilabel performace of the RSPC. For this dataset all 23,149 training samples but only the final 20,000 test samples were used. No further preprocessing or normalization was performed on any of the datasets. 

\subsection*{RST as an unsupervised preprocessing step}
\begin{figure} [H]
\centering
\hspace*{0cm}
\captionsetup{width=17cm}
\includegraphics[trim=0cm 0cm 0cm 0cm,clip,width=14cm]{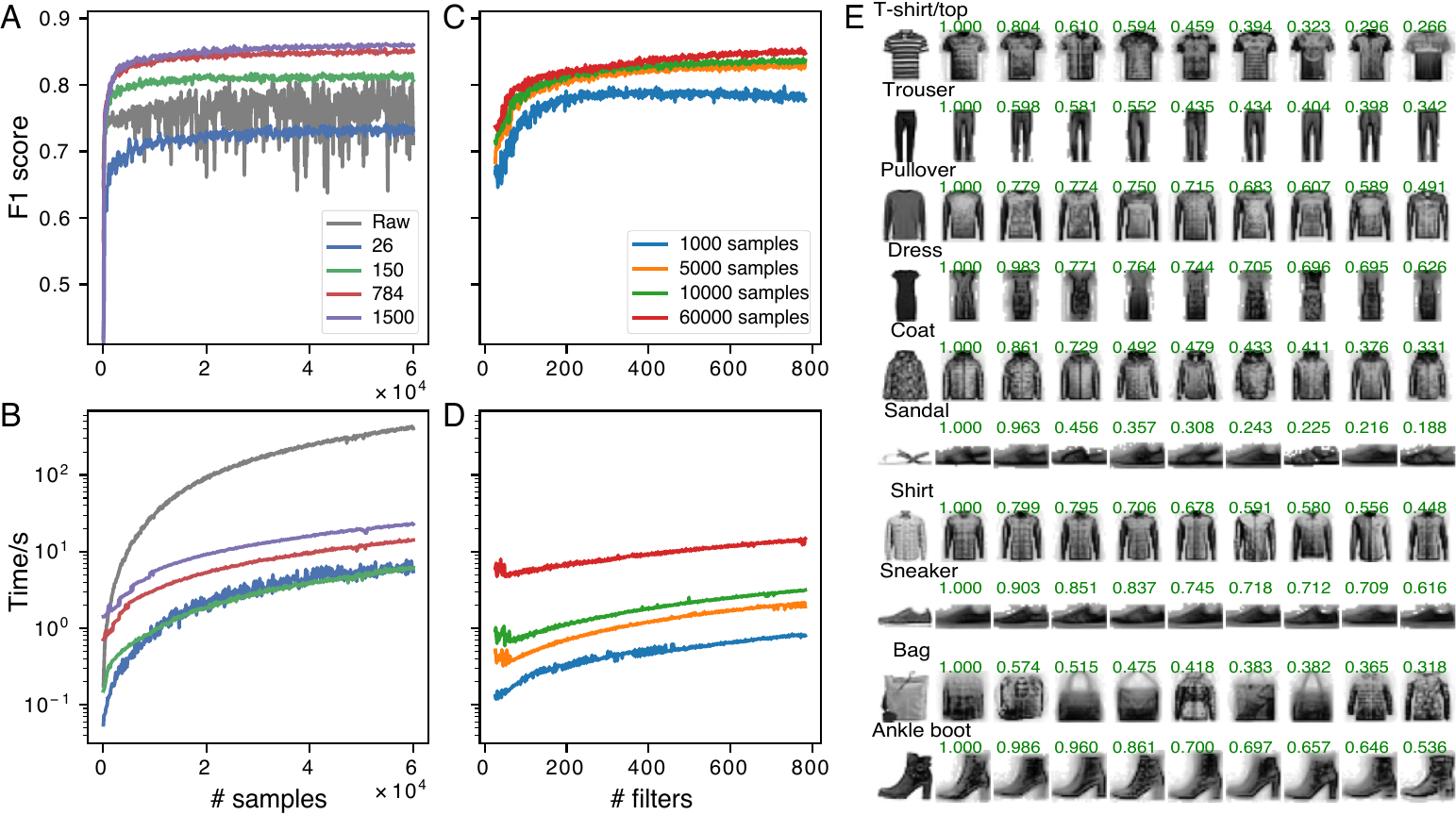}
\caption{\small \textbf{RST as an unsupervised preprocessing step for an SVM A}. Classification performance of SVM classifier on the Fashion-MNIST dataset against number of training examples. Lines are different numbers of filters used to transform the data. Note with 150 filters (150 dimensional input to SVM) performance is already increased compared to the original data (784 dimensions). \textbf{B}. As A but showing duration of transformation and classification. Note that the y axis is logarithmic. \textbf{C}. Classification performance of SVM classifier on the Fashion-MNIST dataset against number of filters. Lines are different numbers of training examples. \textbf{D}. As C but showing duration of transformation and classification. \textbf{E}. Visualization of RSF weights. One random test image was chosen from each class and the weights of the top 8 responding rank filters displayed next to it. In green above the filter weights is their activation in the RST of the image (this value was fed into the SVM as a preprocessing step in A-D).}
\label{fig:RST}
\end{figure}

Linear SVMs are a fast and commonly used classifier that are unable to solve nonlinear problems. To demonstrate the efficiency and utility of RSF we first examine the rank similarity transform (RST) when used as an unsupervised preprocessing step for a linear SVM. For this demonstration we used Fashion-MNIST, and chose parameters for the linear SVM based on the best performance reported on the Fashion-MNIST \href{http://fashion-mnist.s3-website.eu-central-1.amazonaws.com/}{benchmark}. Classification performance for the SVM on the dataset was calculated across different numbers of training samples (see the grey line in Figure \ref{fig:RST}A). As the number of training samples increased, the necessary CPU time also increased strongly (see Figure \ref{fig:RST}B
). When we first transformed the data using RST and then trained a linear SVM with identical parameters the efficiency of the SVM increased. As can be seen in Figure \ref{fig:RST}A, once the number of filters was 150 or greater (compared to 784 original dimensions) this also resulted in an increase in performance and reliability. In addition, it resulted in a dramatic speedup of the SVM classifier of almost 2 orders of magnitude (see Figure \ref{fig:RST}B), although the additional transform process was included in the calculated time. This speedup is not only present when the number of dimensions was reduced but also when it is increased (compare the grey raw data line (d=784) to red (d=1500) line in Figure \ref{fig:RST}B). This is because most dimensions are zeroed after RST, with only the values of the top $k=25$ responding filters are preserved. This was not the case for PCA or KPCA (see Supplementary Figure \ref{fig:PCA}). When dimensions were reduced with PCA there was a slight speedup (with less than $\sim$50 PCs) but a performance decrease (Figure \ref{fig:PCA}A and B). Using KPCA there was neither a speedup nor a performance improvement compared to SVM alone (Figure \ref{fig:PCA}C and D).

\vspace{4mm}

\ni We next looked at the trade off between the number of filters necessary to successfully transform different numbers of training samples and the time taken for training. For low numbers of training samples ($\sim$100 examples per class) increasing the number of filters did not result in much of a performance improvement (blue line in Figure \ref{fig:RST}C) but did result in a longer training time (blue line in Figure \ref{fig:RST}D). However, with higher numbers of training samples ($\sim$6000 per class), adding more filters continues to improve performance (see red line in Figure \ref{fig:RST}C) while also increasing training time (Figure \ref{fig:RST}D). 

\vspace{4mm}

\ni As Fashion-MNIST is an image dataset we were able to visualise the weights of some of the most strongly activated rank filters for 60000 training samples (see Figure \ref{fig:RST}E). Although the RST was unsupervised, filters have been created from similar images resulting in weights that match the test image well. The background blur that is visible for some filters is due to the rank procedure which highlights small changes in the background colour.

\subsection*{Rank similarity classifier}

\begin{figure} [H]
\centering
\hspace*{0cm}
\captionsetup{width=17cm}
\includegraphics[trim=0cm 0cm 0cm 0cm,clip,width=15cm]{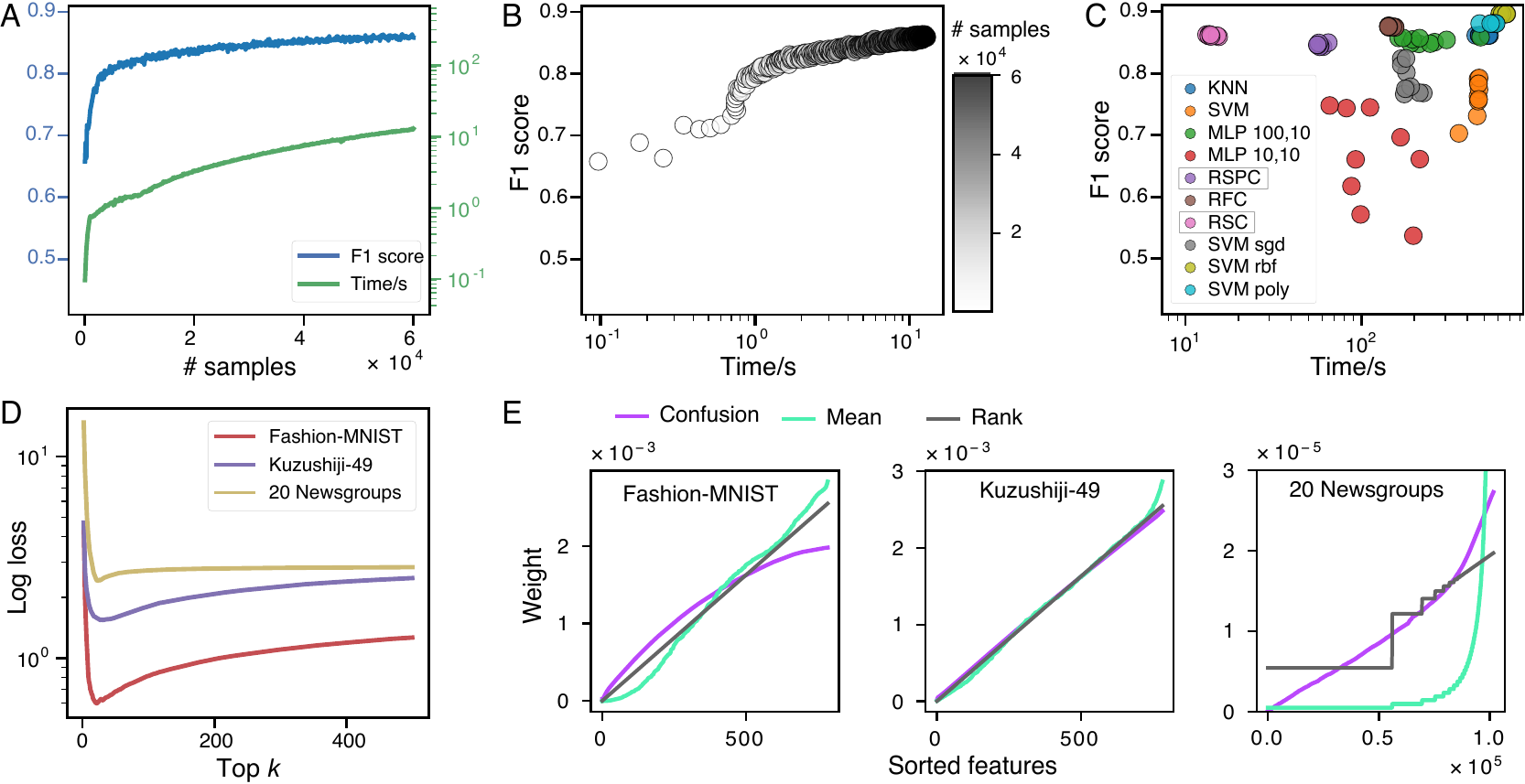}
\caption[]{\small \textbf{RSC classification performance A}. Performance and duration of the RSC classifier depending on number of training samples. Note the increased performance and speed compared to when RST was used as an unsupervised preprocessing step to a SVM in Figure \ref{fig:RST}. \textbf{B}. Same data as A displayed as a scatter plot.  \textbf{C}. Comparison of classifiers' performance and speed on Fashion-MNIST dataset. RSC is on the top left. The parameters of other classifiers were chosen from the best performing or fastest after a previously performed parameter search. \textbf{D}. Accuracy of RSC probabilities. The log loss of the probabilities for three datasets with different values of n\_best. Note that the dataset with the fewest number of classes (Fashion-MNIST) also has the lowest log loss. \textbf{E}. Confusion, mean and rank distributions for sorted features for the Fashion-MNIST, Kuzushiji-49 and 20 Newsgroups datasets. Each feature was averaged and then sorted from small to large according to its mean. The values of the sum of the distribution were normalized to sum to 1. Note that for the 20 Newsgroups dataset the rank distribution diverges strongly from the mean distribution.}
\label{fig:RSC}
\end{figure}

Here we evaluate the performance of rank similarity classifier (RSC) on multiple classification tasks. On the same Fashion-MNIST dataset with increasing numbers of training samples, an RSC with default parameter values outperforms both the pure SVM and the SVM with RST transformed input (compare Figure \ref{fig:RST}A with Figure \ref{fig:RSC}A blue line). Additionally it is more efficient (compare Figure \ref{fig:RST}B with Figure \ref{fig:RSC}A green line), although it is allocating between 1000 and 10000 filters, while this was capped at 1500 for RST. In Figure \ref{fig:RSC}B it can be seen that the F1 score and time taken both increase together as the number of samples increases.

\subsection*{Comparison with other classifiers}

We next examined the performance and efficiency of RSC against the best performing classifiers on the full Fashion-MNIST dataset. We chose the classifiers with the parameters giving the best performance or efficiency reported on the Fashion-MNIST \href{http://fashion-mnist.s3-website.eu-central-1.amazonaws.com/}{benchmark} page.
In Figure \ref{fig:RSC}C it can be seen that RSC is consistently much faster than the other classifiers on this dataset while still producing a comparable classification performance. Note also that the other classifiers were selected after a parameter search (which would multiply the run time by the number of parameters checked) while RSC used default parameters. 

\vspace{4mm}

\ni As can be seen in  Figure \ref{fig:RSC}C, RSC was competitive with two commonly used classifiers (SVC and KNN) in terms of performance on the Fashion-MNIST dataset. Using 10-fold cross-validation (CV) on three datasets, Fashion-MNIST (d=784, c=10), Kuzushiji-49 (d=784, c=49) and 20 Newsgroups (d=10000, c=20), two of images and one of text, we show that this method performs well above chance and gives stable results across multiple datasets (see column 1 of Table \ref{tab:perf}). 

\vspace{4mm}

\ni Also included in Figure \ref{fig:RSC}C is RSPC. Due to needing to calculate the probabilities of each class in the dataset it took longer than RSC, and because the data is not split evenly between the classes it did not perform as well. Despite this, it was more efficient than the other classifiers on this dataset and performed better than the classical SVM or MLP.

\vspace{4mm}
\begin{table} [H]
\begin{tabular}{lrrr}
\toprule
          & RSC (mean\textpm std) & SVM (mean\textpm std) & KNN (mean\textpm std)\\
\midrule
 Fashion-MNIST (10) & 0.8678 \textpm 0.0042 & 0.7765 \textpm 0.0274 & 0.8668 \textpm 0.0034\\
 20 Newsgroups (20) & 0.6120 \textpm 0.0199 & 0.6424 \textpm 0.0150 & 0.2290 \textpm 0.0105\\
  Kuzushiji-49 (49) & 0.9073 \textpm 0.0151 & 0.4303 \textpm 0.0333 & 0.8771 \textpm 0.0211\\
\bottomrule
\end{tabular}
\caption{\small Table with average results from 10-fold CV of the full datasets (test and training data combined) for 3 classifiers.}
\label{tab:perf}
\end{table}

\subsection*{Probabilities}
Probablilites for RSC for the three datasets from Table \ref{tab:perf} were calculated as described in the Methods(Algorithm \ref{alg:prob}). We used log loss (cross-entropy loss) to evaluate the accuracy of the probabilities on test data in Figure \ref{fig:RSC}D. This value was acceptably low for the Fashion-MNIST dataset, although the probabilities were not learned or adjusted to the data at all. For the Kuzushiji-49 and 20 Newsgroups datasets the log loss was higher, which may reflect the lower accuracy of the probabilities for those datasets but was also due to the increased number of classes (c=49 and c=20 respectively). As seen in Figure \ref{fig:RSC}D accuracy of the probabilities was dependent on the value of the hyperparameter $k$ that defines $S_Imin$ (see Algorithm \ref{alg:prob}). This is because only the top $k$ filters have any effect on the probabilities. When $k$ was too low, relevant classes were not included in the calculation. When $k$ was too high, irrelevant classes had non-zero probabilities which added noise.

\subsection*{Filter distribution on datasets}

Here we compare the calculated confusion distribution with the ranks of features on the three examined datasets. While ranks were a good approximation for both the Fashion-MNIST and Kuzushiji-49 datasets, for the 20 Newsgroups they are more divergent (Figure \ref{fig:RSC}E, compare to Figure \ref{fig:confusion}E). This is due to the log-log increase in word frequencies with the rank its frequency of use in natural language (as described by "Zipf's law"\cite{Powers1998-dm}). When the ranks were replaced by confusion distribution values for 20 Newsgroups there was a 4.6\% improvement in the F1 score (0.523$\pm$0.021 to 0.547$\pm$0.031).

\subsection*{Rank similarity probabilistic classifier}

\begin{figure} [H]
\centering
\hspace*{0cm}
\captionsetup{width=17cm}
\includegraphics[trim=0cm 0cm 0cm 0cm,clip,width=14cm]{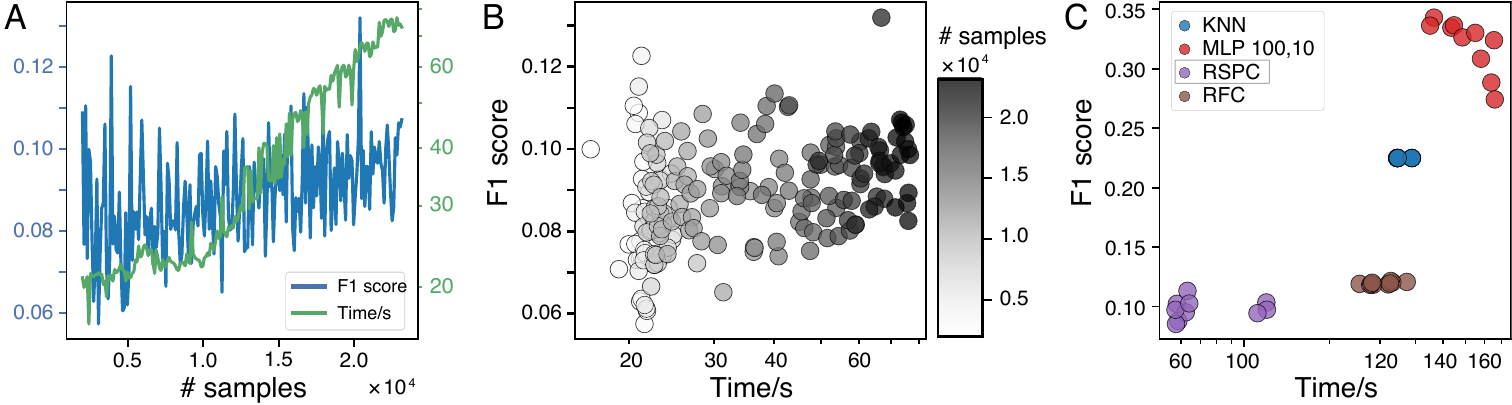}
\caption{\small \textbf{Multilabel classification A}. Performance and efficiency of RSPC on RCV1 dataset (d=47236, c=103) with variable numbers of training samples. \textbf{B}. Same data as A displayed as a scatter plot. \textbf{C}. Comparison of classifiers on RCV1 dataset, RSPC is on the bottom left. Despite poor performance on this dataset, RSPC is still above chance and faster than the alternatives that were able to complete the classification task using only 8Gb of RAM.}
\label{fig:multilabel}
\end{figure}

Since many large datasets are both multiclass and multilabel it is useful to have a classifier that can handle this. While RSC is not natively suitable for the multilabel case, RSPC is able to handle this situation. Here we showed that on a subset of the commonly used text classification dataset RCV1 (d=47236 sparse, c=101), RSPC is able to perform well above chance (see Figure \ref{fig:multilabel}A and B). The subset used here was the full 23149 training samples but only the final 20000 test samples. When compared with other classifiers on this dataset it does not perform as well (see Figure \ref{fig:multilabel}C) however, it is still much faster than the alternatives and performs well above chance. When used on the more suitable Fashion-MNIST dataset it was able to perform competitively with the other classifiers (see Figure \ref{fig:RSC}C).

\section*{Discussion}

Here we demonstrated that RSFs are a quickly converging nonlinear ANN that offers competitive performance on benchmark datasets with a wide array of structures and statistics. With our insight of viewing features as distributions to be discriminated, the use of ranks can be supplemented by a confusion measure. Our software package allows the RSFs to be used as a drop in replacement for scikit-learn transformers and classifiers. This showed RSC to be computationally efficient and almost an order of magnitude faster than other scikit-learn classifiers.

\vspace{4mm}

\ni RSC has two other main advantages in comparison with other classifiers. Firstly, by using prototypical vectors the model is intuitively understandable (see for example \ref{fig:RST}E). This is a desirable feature to protect against overfitting or flaws in uninterpretable "black box" models like DNNs \cite{Rudin2019-wv}. Secondly, by using ranks of features we utilise many of the advantages of nonparametric statistics \cite{Conover1981-us}, namely that we do not need to make an assumption about the distribution of the underlying data, and that the filters are scale invariant. This means that data preprocessing, which is used extensively in machine learning, will often not be necessary for RSF thereby increasing its efficiency and ease-of-use.  

\vspace{4mm}

\ni The RSFs are L1 normed, since ranked data with the same number of features are L1 equal. We kept this equal L1 distance (Manhattan distance) so that the values are a measure of how many ranks different each individual feature is; if all features are equal then the activation strength of all RSFs are equal. 
Homeostatic plasticity appears to be an important feature of biological neurons, where balanced changes in synaptic weights \cite{Royer2003_hp} and membrane conductances \cite{Turrigiano2004_hp}, as well as the intrinsic properties of neuronal dendrites \cite{Hausser2001_dd}, help to maintain stable input-output relationships over time and potentially improve computational performance \cite{Bird2021-iq}. A recent paper on synaptic weights measured in cortical neurons \textit{in vivo} shows that the average weight across synapses is constant across time \cite{Melander2021-av}, which is specifically equivalent to maintaining an equal L1 distance as here.

\vspace{4mm}

\ni While there are many advantages to this method, one drawback is that it is unsuitable for very low dimensional data. As $n$ features have only $n!$ possible combinations (not accounting for ties) the number of points should be much lower than $n!$ for the prototypes to represent the space. Therefore, for most practical applications at least 9 features are necessary, which allows for a minimum of 362,880 unique filters. RSFs are also not suitable for data where specific features take a different form to others and are much larger or smaller, although within feature normalization can be performed before application to take care of this issue. Using RSF is most suitable when features have the same units and are part of the same space, for example, pixels in the same image or word counts in a text document. If the features are unrelated to each other then the order statistics may no longer be meaningful, although separate features may be normalized to the same scale.

\vspace{4mm}

\ni In future work, this algorithm could be sped up further by parallelization via GPU or neuromorphic computing. Neuromorphic chips are highly parallel and energy efficent, designed to imitate the brain \cite{Schuman2017-hb}. They have been been used to solve a wide range of problems \cite{Costas-Santos2007-oa, Blum2017-qt}, and recently millions of neurons have been connected to implement an efficient KNN search \cite{Paxon_Frady2020-gn}. Our algorithm could be similarly implemented on such a chip as it is also dot product based, winner-takes-all, and has the added benefit of natively using integer weights. As a VQ method it could additionally be extended to problems like data compression or dimensionality reduction. 

\vspace{4mm}

\ni Overall, RSF performs as well as similar methods and (even without parallelisation) far more efficiently. This makes it a useful alternative when interpretable and efficient models are needed.

\section*{Acknowledgements}
We would like to thank Prof. Dr. Wolf Singer and Dr. Felix Effenberger for their comments on this manuscript. 
We acknowledge funding through the grant of Prof. Dr. Wolf Singer from DFG Reinhart Koselleck (project number 325248489) and support from the Ernst Strüngmann Institute (ESI) for Neuroscience in Cooperation with Max Planck Society.


\section*{Supplementary Material}

\paragraph*{Code:} 
Our code is available here: \url{https://github.com/KatharineShapcott/rank-similarity} 

\paragraph*{Supplementary Figures.} \ref{fig:PCA} 


%
%
%

\bibliographystyle{hunsrtnat}
\bibliography{RankSimilarity}

\FloatBarrier
 \setcounter{table}{0}
\renewcommand{\thetable}{S\arabic{table}}
\renewcommand{\thefigure}{S\arabic{figure}}
\setcounter{figure}{0}

\section*{Supplementary Figures}\label{S1_File}
\begin{figure} [H]
\centering
\hspace*{0cm}
\captionsetup{width=17cm}
\includegraphics[trim=0cm 0cm 0cm 0cm,clip,width=14cm]{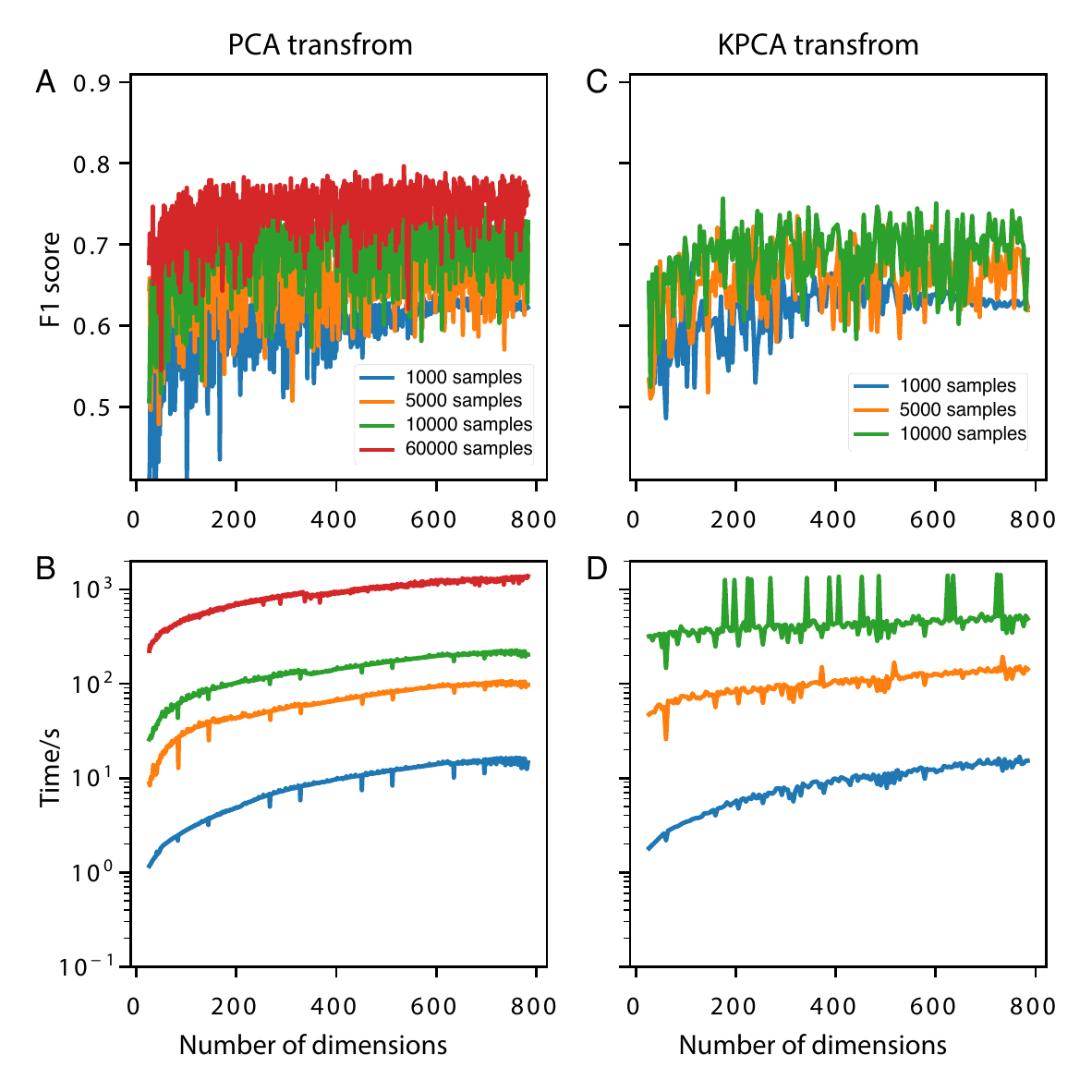}
\caption{\small \textbf{Comparison with dimensionality reduction methods. A}. Classification performance of SVM classifier on the Fashion-MNIST dataset after PCA dimensionality reduction. Lines are numbers of samples. There is no improvement in performance. \textbf{B}. As A but showing duration of transformation and classification. Note that there is little speed improvement compared to using RST in Figure \ref{fig:RST}D. \textbf{C}. Classification performance of SVM classifier on the Fashion-MNIST dataset after nonlinear KPCA dimensionality reduction. Lines are numbers of samples. The performance is worse than RST and similar to PCA. Note that using 8Gb of RAM KPCA was not able to transform the full training dataset of 60000 images. \textbf{D}. As C but showing duration of transformation and classification. Using KPCA takes even longer than PCA. Note that using 8Gb of RAM KPCA was not able to transform the full training dataset of 60000 images.}
\label{fig:PCA}
\end{figure}

\end{document}